\documentclass{article}
\usepackage{ijcai18}

\usepackage{times}
\usepackage{xcolor}
\usepackage{soul}
\usepackage[utf8]{inputenc}
\usepackage[small]{caption}

\usepackage{amssymb}
\usepackage{booktabs}
\usepackage{graphicx}
\usepackage{subfigure}
\usepackage{rotating}
\usepackage{array}

\title{Learning Transferable UAV for Forest Visual Perception}

\author{
Lyujie Chen,
Wufan Wang,
Jihong Zhu
\\
Beijing National Research Center for Information Science and Technology (BNRist) \\
Department of Computer Science and Technology, Tsinghua University, Bejing, China\\
\{chenlj16, wwf14, jhzhu\}@mails.tsinghua.edu.cn
}

\begin{document}

\maketitle

\begin{abstract}

In this paper, we propose a new pipeline of training a monocular UAV to fly a collision-free trajectory along the dense forest trail. As gathering high-precision images in the real world is expensive and the off-the-shelf dataset has some deficiencies, we collect a new dense forest trail dataset in a variety of simulated environment in Unreal Engine. Then we formulate visual perception of forests as a classification problem. A ResNet-18 model is trained to decide the moving direction frame by frame. To transfer the learned strategy to the real world, we construct a ResNet-18 adaptation model via multi-kernel maximum mean discrepancies to leverage the relevant labelled data and alleviate the discrepancy between simulated and real environment. Simulation and real-world flight with a variety of appearance and environment changes are both tested. The ResNet-18 adaptation and its variant model achieve the best result of 84.08\% accuracy in reality.

\end{abstract}

\section*{Videos and Dataset}

Additional videos and the full training/testing datasets are available at https://sites.google.com/view/forest-trail-dataset.

\section{Introduction}
Unmanned Aerial Vehicles (UAVs) have been increasingly popular in many applications, such as search and rescue, inspection, monitoring, mapping and goods delivery. For UAV with very limited payloads, visual technics provide a more feasible way to perceive the world instead of state-of-art radars. In this paper, we primarily study the problem of navigating a monocular UAV in the dense forest by finding a collision-free trajectory, which simultaneously follows the trail and avoids the obstacles.

In recent years, learning based method exceeded the traditional hand-engineered perception and control method. However, because of the wide appearance variability and task complexity, features of clustered forest are much more difficult to extract and learn. As a result, a large amount of data is required to use supervised deep learning.
While there has been some progress in navigating drone in forest trail \cite{giusti2016machine}, the image dataset provided has so much man-made deviation which leads to wrong image label. Real-time images of the clustered forest with attitude information are expensive and difficult to collect by autonomous navigation. A tiny mistake will lead to the catastrophic and dangerous result. Therefore, acquiring more easily obtained data from simulation environment is an effective alternative.
But visual appearance between simulated and real world is not the same. In traditional machine learning method, training and test data are forced to have the same data distribution and input feature space. The learned policies are only put to use in the similar environment and domain that the model was originally trained on. In order to apply policies from simulation to real flight, it is essential to use transfer learning to leverage the relevant labelled data and alleviate the discrepancy between different environments. In general, forest perception and autonomous navigation in the clustered forest is still a challenging task.

\begin{figure}[]
\centering
\includegraphics[width=\linewidth]{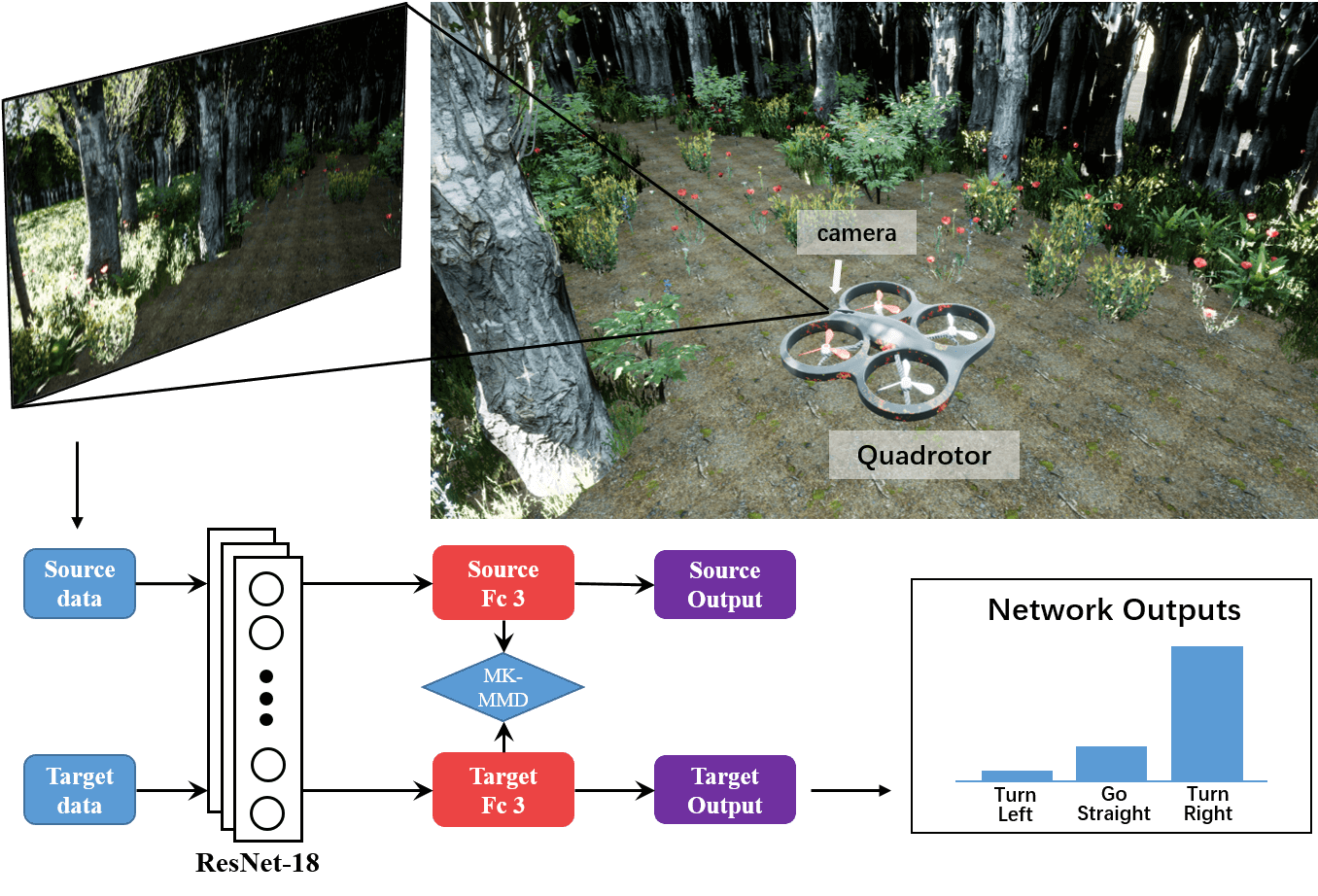}
\caption{A quadrotor acquires the forest images from a forward-looking camera;
a ResNet-18 adaptation network outputs the probabilities of three defined classes, which will be transformed into the control signal of UAV by a simple reactive controller.}
\label{fig1}
\end{figure}
%

\begin{figure*}[]
\centering
\subfigure[Scenes in Trail 1]{
\includegraphics[width=0.95\textwidth]{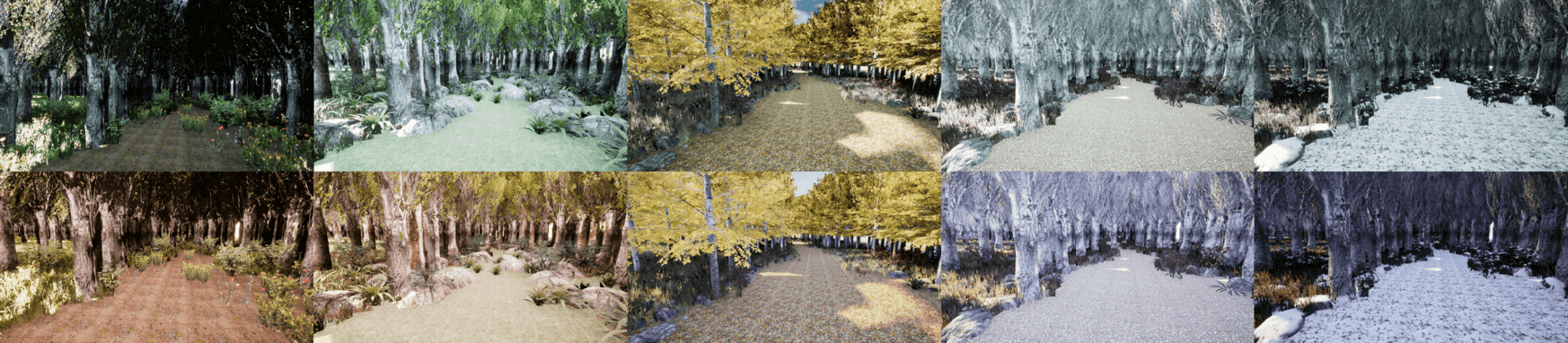}
}
\subfigure[Scenes in Trail 2]{
\includegraphics[width=0.95\textwidth]{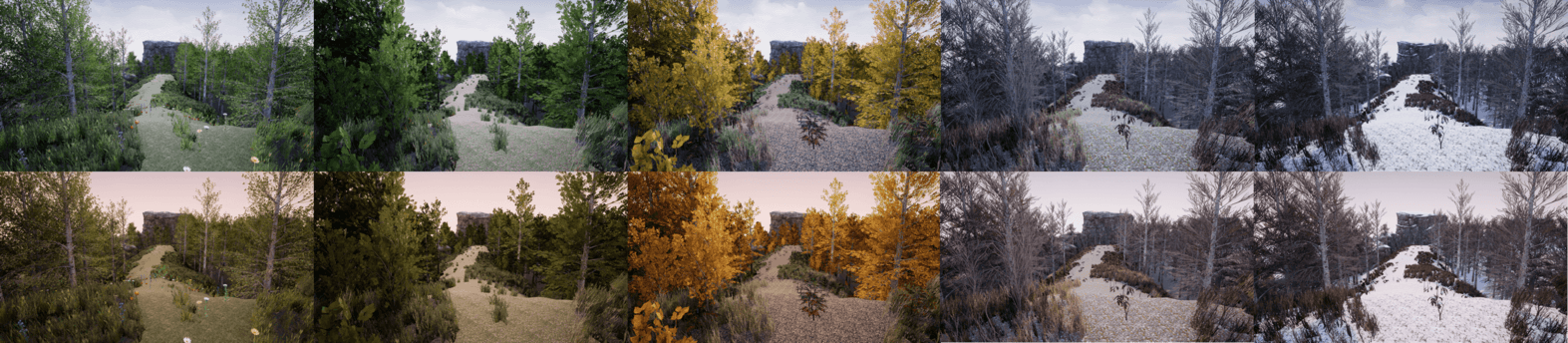}
}
\caption{Simulated forest trail dataset. From the first column to the last column represent the forest scenes of spring, summer, autumn, winter and snow respectively. For each trail, the first row is the forest in the morning and the second row is at dusk.}
\label{fig2}
\end{figure*}

In this paper, we solve the problem by first collecting a new dense forest trail dataset in the simulation. Collecting data from the real world is expensive and time-consuming. To retrieve most similar and real pictures of natural scenery, we utilize a new simulator named Airsim \cite{airsim2017fsr} which offers physically and visually realistic simulations built on Unreal Engine to gather a large number of annotated training data in a variety of seasonal conditions and terrain environments.
Based on a large amount of data, a deep neural adaptation network is constructed to learn a UAV navigation strategy step by step. Visual perception of forests is viewed as a classification problem. During the flight, proposed model predicts the moving direction to keep the UAV remain on the trail.

The proposed method is validated by a series of experiments, where simulation and real-world flight with a variety of appearance and environment changes are both tested. Among them, we compare the influence of season, lighting, terrain and domain similarity on the effect of transferring improvement. In all tasks, the adaptation model achieves a much better result. It can be observed that different source data affects the adaptation performance to varying degrees. Also, the larger differences between domains, the worse performance achieves on basic model and the greater performance boost on adaptation model. In addition, we propose a new multi-source adaptation model to validate the observations, which gives a very good intuition and a general guideline for making full use of source training data in transfer learning.

The contributions of our work are three folds. (1) A dense forest trail image dataset in a variety of environments. (2) A forest visual perception technique based on deep neural network. (3) A transfer learning technique to control UAV in real world based on the policies learned in simulation environment.

\section{Related Work}
\subsection{Vision Based Forest Perception on UAVs}
While Laser rangefinders usually only support 2D detections and radars are too heavy for flight, lightweight vision sensors become a more effective and reliable alternative to perceive the forest. Vision-based techniques on small outdoor UAVs are widely used for obstacle avoidance and path tracking \cite{giusti2016machine,ross2013learning,daftry2016robust,dey2016vision,barry2015pushbroom}.

Monocular vision techniques are most widely used and attractive because they only rely on a single available camera which is lightweight and easy to deploy. The monocular image can be used as direct input of machine learning \cite{giusti2016machine} and imitation learning method \cite{ross2013learning}. It also can be presented as a depth estimation \cite{daftry2016robust,dey2016vision}.

Stereo systems are always used to compute optical flow and depth map \cite{byrne2006stereo,yang2003multi}. With more data to be processed, optimization algorithm becomes the crucial part for the real-time flight of stereo vision UAV. Recent work \cite{barry2015pushbroom} performs an optimized block-matching stereo method to fly a small UAV at over 20 MPH near obstacles and is able to detect the obstacle at 120FPS on an airborne CPU processor.


With the development of artificial intelligent technology, deep learning based methods are widely used in the field of the unmanned system. \cite{levine2016end} regard the robot control problem as a supervised training process and use a guided policy search to map raw image to robot's motors signals directly. \cite{giusti2016machine} propose a monocular UAV to perceive and follow the forest trail based on a deep neural network classifier. It manually collected about 100K images by hiking approximately 7 kilometres forest and mountain trail.

\begin{figure}[]
\centering
\includegraphics[width=0.9\linewidth]{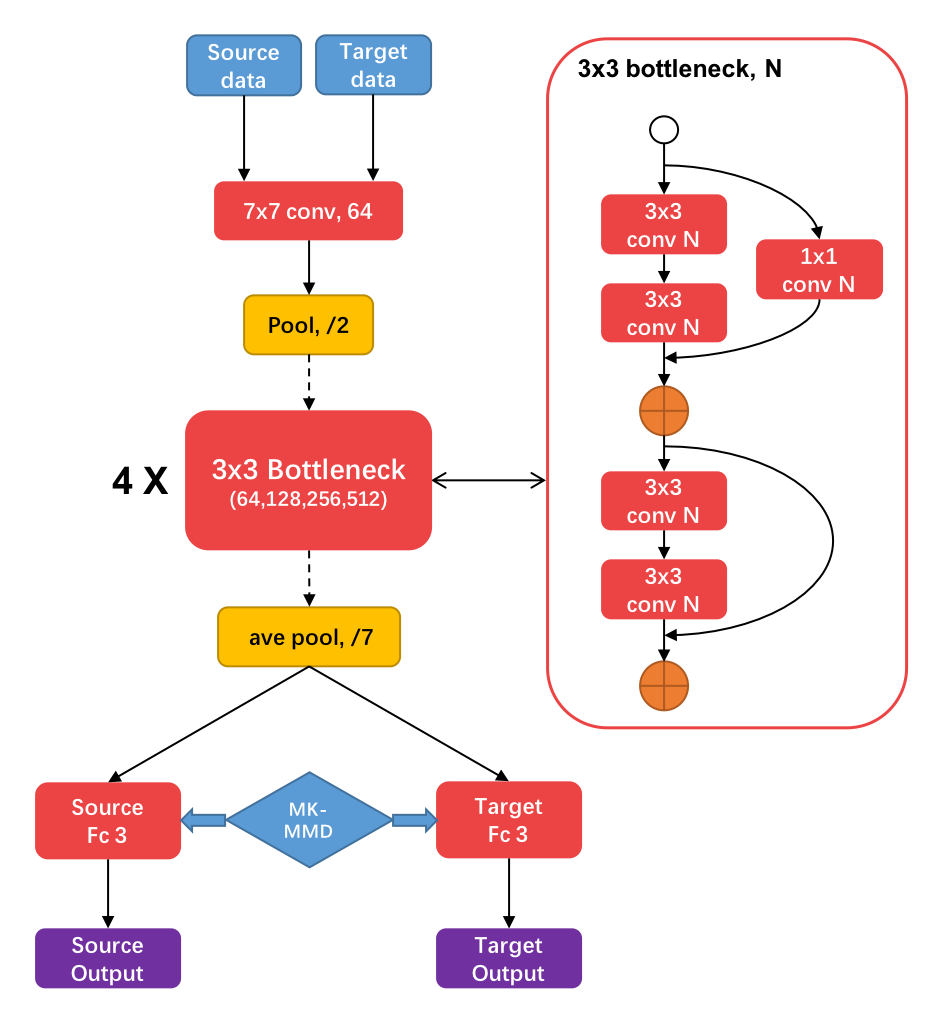}
\caption{ResNet-18 adaptation network. Convolutional kernel size in all bottlenecks is 3 $\times$ 3 while the filter number goes up from 64 to 512 with every doubling. All convolutional layers are trained via standard SGD. The last fully connected layer is trained to classify the images in the specific domain and should be adapted by MK-MMD.}
\label{fig3}
\end{figure}

\subsection{Transfer Learning}
Transfer learning focuses on applying knowledge gained from solved problems to a different but related problem \cite{weiss2016survey}. There are two common ways of transfer learning in the field of deep learning. The most commonly used one is fine-tuning. Many deep learning based methods fine tune from the existing trained model because the target task doesn't have enough data. The other way of transferring knowledge is to change the structure of neural network \cite{long2015learning,long2016deep}. In this paper, we combine these two methods, which fine tune from a basic learned model while following the deep adaptation network \cite{long2015learning} approach to enhance the ability of autonomous flight in reality.

Deep adaptation network \cite{long2015learning} computes domain discrepancy as the mean embedding of all task-specific layers and adds them to whole network loss. The model jointly improves the performance of source and target. \cite{daftry2016learning} utilize deep adaptation network for autonomous MAV flight using monocular reactive control. They use Zurich forest trail dataset \cite{giusti2016machine} as source domain and evaluate the effectiveness of proposed method through real-world flight experiments.

Compare with previous works, we gather a more accurate and comprehensive dataset and train a more robust CNN model based on modern transfer learning method.

\section{Learning Transferable UAV}
In this section, an adaptation network is constructed step by step. We first explain the reason for collecting new dataset and elaborate on gathering a large and representative labelled forest dataset both in the simulated and real-world environment. Then we adopt a ResNet-18 \cite{he2016deep} model to perceive the forest by deciding the moving direction for given simulated images. Since the real world has the very different visual appearance, the model learned from simulation hardly performs well in reality. Hence, in order to transfer the learned policies, we further introduce a cross-domain discrepancy metric and propose a new adaptation network. Unlabeled images in target domain are used to enhance the ability of autonomous flight in reality.

\subsection{DataSet}

In this paper, we do not use the off-the-shelf Zurich forest trail dataset \cite{giusti2016machine} because of the man-made deviation problem. All the images in that dataset were collected by three head-mounted cameras equipped on a hiker. Although the author claimed that they took care of always looking straight along its direction of motion, there were still some destructive and meaningless head-turning and shaking. What's more, the hiker sometimes hesitated when facing sharp turn and not headed straight along the trail. Both of them produced a lot of wrong annotated data.

Collecting data from the real world is difficult and time-consuming. Therefore, there is a need to acquire more easily obtained data from simulation environment. To narrow the gap between simulation and reality, we build variable conditions and environments in Unreal Engine which generate reasonably realistic reconstruction of real-world scenes. We use a simulator called Airsim \cite{airsim2017fsr}  to offer a quadrotor for gathering the images.
To carefully follow the dominant direction of the trail and gather high-precision images, we start to make the UAV fly a smooth trajectory in a realistic simulation environment. Next, we walk through the real forest roads by holding the cameras instead of flying a real UAV. We refer to the same acquisition way of \cite{giusti2016machine} by mounting three cameras for easily labelling the data. One pointing straight ahead, and the other two pointing 30 degrees to the left and right respectively.
All images acquired by three cameras are labelled. We define three classes, which correspond to three actions that UAV should take to remain on the trail. Specifically, the central camera acquires instances for the class GS(Go Straight). Conversely, all images acquired by the right view camera are of TL(Turn Left) class; and all images acquired by the left-looking camera are of TR(Turn Right) class.

Our forest dataset consists of simulated and realistic parts. The former is composed of 99762 images spread over four seasons, two different trails and several different light conditions as well as viewpoint heights (See Fig.\ref{fig2}). The latter is a set of 11103 images composed by 1 hour of 1920 x 1080 30fps video acquired using three portable cameras.

\begin{table*}[]
\centering
\resizebox{\textwidth}{!}{%
\begin{tabular}{@{}ccccccc@{}}
\toprule
Task & Training Data Source & \begin{tabular}[c]{@{}c@{}}Number of \\ Training Data\end{tabular} & Validation Data Source & \begin{tabular}[c]{@{}c@{}}Number of \\ Validation Data\end{tabular} & Test Data Source & \begin{tabular}[c]{@{}c@{}}Number of \\ Test Data\end{tabular} \\ \midrule
Transfer across season condition & Trail 1 without Winter & 41814 & Winter of trail 1 & 300 & Winter of trail 1 & 28005 \\
Transfer across terrain condition & All season of trail 1 & 70419 & All season of trail 2 & 300 & All season of trail 2 & 29043 \\
Transfer across light condition & All morning data & 49752 & All evening data & 1000 & All evening data & 49010 \\ \bottomrule
\end{tabular}
}
\caption{Transfer Learning Tasks in Simulation Dataset}\label{tb1}
\end{table*}

\subsection{Deep Neural Network for Forest Perception}

Convolutional neural network (CNN) has been shown to perform well in classification problems given a large number of images  \cite{he2016deep,szegedy2015going,szegedy2017inception}. The problem of finding a collision-free trajectory in the forest can be formulated as the problem of classifying the control action of UAV frame by frame.

In this paper, we adopt a CNN as an image classifier based on ResNet structure. It consists of successive pairs of convolutional and batch normalization bottlenecks, followed by a fully connected layer.
In particular, the input layer is considered as a matrix of 3 $\times$ 224 $\times$ 224 neurons. The input image is resized to 224 $\times$ 224 before mapped to the neurons in the input layer. For a given input, the CNN outputs three values, indicating the probability that the UAV will turn left, turn right and go straight respectively.
The training set is only augmented by the horizontal flip to Synthesize the left/right mirrored images. A mirrored training image of class TR (TL) produces a new training sample for class TL (TR). A mirrored training image of class GS still yields a new sample of class GS. Augmentation has double the number of samples.

The model is implemented in Caffe \cite{jia2014caffe} and trained using standard backpropagation. Weight parameters $\{W^l\}_{l=1}^l$ in convolutional layers are initialized with MSRA filter \cite{he2015delving} and all bias parameters $\{b^l\}_{l=1}^l$ are initialized to 0. All parameters are jointly optimized using SGD to minimize the misclassification error over the training set (See Eq.\ref{cnn}).

\begin{equation}
    \min_{\Omega} {\frac{1}{n} \sum_{i=1}^{n}{J(\theta(x_i),y_i)}} \label{cnn}
\end{equation}

where $\Omega=\{W^l, b^l\}_{l=1}^l$ denotes the set of all CNN parameters among $l$ layers. $n$ is the number of training data. $J$ is the cross-entropy loss function. $\theta(x_i)$ is the output probability of CNN and $y_i$ is the ground truth label given input $x_i$.

In the test phase, an image from the monocular camera will be fed into the trained model. The output probability corresponds to the control signal of UAV. We implement the same simple reactive controller as \cite{giusti2016machine}, which only control the yaw rate and velocity of the flight. The desired velocity is proportional to the probability of GS. The desired yaw rate is proportional to the probability difference between TR and TL, which is P(TR)-P(TL). When the value is positive, UAV is steered to the right. Conversely, a negative value steers the flight to the left.

\subsection{Transferable Policy Using Deep Adaptation Network}

So far, the learned strategy can only be applied to the simulation environment. However, the real world has very different and variable appearance from the simulation. In this section, we study a problem of transferring our learned policies to real-world flight.

According to the definition of transfer learning, all datasets are divided into two domains. In our problem, the source domain is simulation environment while the target domain is real world, which are characterized by probability distributions $p$ and $q$, respectively. All data in source domain is labelled, which can be denoted as $\emph{D}_s = \{(x_i^s, y_i^s)\}_{i=1}^{n_s}$ with $n_s$ samples. All data in target domain is unlabelled, denoted as $\emph{D}_t = \{(x_j^t)\}_{j=1}^{n_t}$ with $n_t$ samples.

In this section, we aim to construct a new deep adaptation network based on the previous ResNet-18 model to bridge the cross-domain discrepancy. The new classifier $y=\theta(x)$ uses the source supervision to minimize the error risk in target domain $ \varepsilon_t(\theta) = Pr_{(x, y)\sim q}[\theta(x) \neq y] $.

\subsubsection{MK-MMD}

Real-world data is hard to collect, which means we have no (or very limited) labelled information in the target domain. To approach this very common challenging problem in domain adaptation, many existing methods aim to simultaneously optimize the performance of source and target domains by introducing a discrepancy metric. In our paper, we use the same measure of domain discrepancy as \cite{long2015learning}, which apply the multiple kernel variant of maximum mean discrepancies (MK-MMD) proposed by \cite{gretton2012kernel}. The proposed measurement focus on jointly maximize the two-sample test ability while minimizing the Type II error, i.e., incorrectly retaining a false null hypothesis.

Given the domain probability distributions $p$ and $q$, the MK-MMD $d_k(p, q)$ is defined as the distance between the mean embedding of $p$ and $q$ in reproducing kernel Hilbert space (RKHS).
The squared formulation of MK-MMD is denoted by

\begin{equation}
    d_k^2(p, q) \triangleq \left \| E_p[\phi (x^s)] - E_q[\phi (x^t)] \right \|_{H_{k}}^2 \label{eq2}
\end{equation}

Where $H_k$ is the reproducing kernel Hilbert space (RKHS) with a characteristic kernel $k$ which correlated with the feature map $\phi$. $E_p[\phi(x^s)]=<\phi(x^s),\mu_k(p)>_{H_k}$
where $\mu_k(p)$ is the mean embedding of distribution $p$ in $H_k$. The most important property of $d_k(p,q)$ is that $p = q \iff d_k^2(p,q) = 0$.

In order to simultaneously minimize the misclassification error and discrepancy between domains, we add an MK-MMD based adaptation regularizer to the fully connected layers of CNN, which introduce the cross-domain discrepancy to basic misclassification error loss (See Eq.\ref{eq3}).

\begin{equation}
    \min_{\Omega} {\frac{1}{n_{s}} \sum_{i=1}^{n_{s}}{J(\theta(x_i^s),y_i^s)}} \label{eq1} +\lambda \sum_{l=l_1}^{l_2} d_k^2(\emph {D}_s^l, \emph {D}_t^l) \label{eq3}
\end{equation}

\begin{figure*}[]
\centering
\subfigure[Six frames in simulation]{
\includegraphics[width=0.95\textwidth]{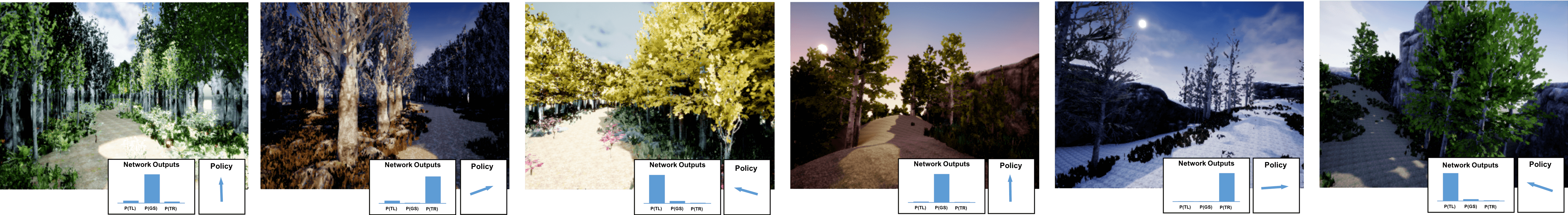}
}
\subfigure[Six frames in real world]{
\includegraphics[width=0.95\textwidth]{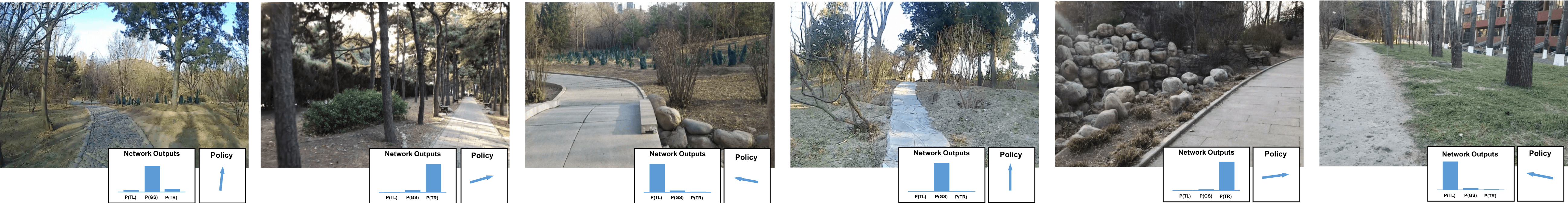}
}
\caption{Six representative frames in simulation and reality. For each frame,  the network outputs and the motion policy are reported based on the ResNet-18 adaptation network.}
\label{fig5}
\end{figure*}

where $\lambda$ is a penalty parameter greater than 0, $l_1$ and $l_2$ are layer bounds between which the regularizer become effective. In our paper, we set $l_1$=17 and $l_2$=18, which is the last fully connected layer. Since features always transit from general to specific along the network, the last fully connected layer in our model is trained to classify the images in the specific domain. Hence, the features are not transferable and need to be adapted with MK-MMD.

The new ResNet-18 adaptation network (See Fig.\ref{fig3}) is still trained using a mini-batch supervised gradient descent with the above optimization framework (Eq.\ref{eq3}). Source and target data are both sent to input layer at the same time. With a large number of unlabeled real-world images, more features of target domain are considered during the training, which will greatly boost the performance.

\section{Experiments}
\subsection{Setup}
In this section, we present experiments to analyze the performance of ResNet-18 adaptation network with MK-MMD layer (ResNet-18-Adap) comparing to the basic ResNet-18 network without transfer learning. In order to evaluate the improvement of transfer learning on the tasks, all the experiments are conducted both on ResNet-18 and ResNet-18 adaptation networks.
We build three learning tasks on simulation dataset (See table \ref{tb1}) to evaluate the proposed methods. These are season change test, terrain change test and light change test respectively. Note that, all validation data is randomly retrieved from all data in the target domain. In addition, a learning task of transferring the policy from simulation to reality is built. We compare the effect of different source data on domain adaptation performance.

In all tasks, we first only use data in the source domain to train a ResNet-18 model from vanilla to expert, applying mini-batch SGD with 0.9 momentum and the learning rate annealing strategy. The initial learning rate is set to be 0.05. It requires about 5 hours on a server equipped with an NVIDIA Titan X GPU. Then, we introduce the unlabeled data in target domain to train a ResNet-18 adaptation network. We set its learning rate to be 0.003 and use the SGD with 0.75 momentum.
After every 300 iterations of training, a test will be conducted on validation set. We only save the model with the best performance on validation set, then finally evaluate on test dataset.

\begin{table}[]
\centering
\resizebox{\linewidth}{!}{%
\begin{tabular}{@{}ccc@{}}
\toprule
Transfer Tasks & ResNet-18 & ResNet-18-Adap \\ \midrule
Transfer across season condition & 71.63\% & \textbf{83.75\%} \\
Transfer across terrain condition & 84.25\% & \textbf{91.33\%} \\
Transfer across light condition & 93.23\% & \textbf{94.33\%} \\
Transfer across simulation and reality & 72.24\% & \textbf{81.40\%} \\ \bottomrule
\end{tabular}
}
\caption{Adaptation Results on All Tasks}\label{tb2}
\end{table}

\subsection{Results and Analysis}
The supervised learning and unsupervised adaptation results on all tasks are shown in Table \ref{tb2}. We compare the accuracy of predicting direction with and without adaptation. The performance boost on all tasks indicates that the architecture of MK-MMD adaptation has the ability to transfer learned policies across source and target domains. After that, we apply the trained adaptation model to a UAV in the simulated world. The test video can be found here: https://sites.google.com/view/forest-trail-dataset.

\subsubsection{Transfer across simulated environments}
In these three experiments, we try to transfer policies over the different simulated environment. The domain shift of season change is induced by the difference in the visual appearance of foliage. While the spring and summer environment is cluttered with dense foliage and the autumn environment has different foliage colour, the characteristics of winter condition are absence of foliage and presence of snow. In this case, the accuracy boost from 71.63\% to 83.75\%. In the scenario of terrain change, the domain shift is mainly induced by the difference in flight altitude and bumpy trail roads. Comparing to season change, this domain difference is smaller. The basic ResNet model has achieved the great result. The adaptation model reaches 91.33\% accuracy rate compared to 84.25\% baseline. The domain difference of light change is the smallest. All plants at dusk immerse in the warmer sunshine and become yellower. But the other characters are the same. The adaptation model only improves about 1\% accuracy rate.

\subsubsection{Transfer across simulation and reality}
During the reality test, we compare the adaptation performance of four different source domain datasets. At first, we use all simulated data to train a basic ResNet-18 model to mak full use of every feature from all seasons and environments. Then unlabeled real-world data is used to train a ResNet-18 adaptation model. The accuracy boost from 72.24\% to 81.40\%. Further, we experiment with three more different subsets of all source images in order to figure out which is beneficial for learning (See Table \ref{tb3}). Since the real world data is gathered in winter, we choose all simulated winter images as the first sub-dataset, trying to extract most of the features in one season. However, the model has very poor performance because the real world forests consist of diverse foliage and terrain compared to clear season appearance difference in the simulated environment. Both evergreen pines and deciduous plants live at the same place. So we choose the second sub-dataset by adding images of spring and summer. At this time, the model has the better result but still worse than the original one. At last, we generate the third sub-dataset as all morning images because the real world dataset has no images at dusk. In this case, the basic model performs better, which demonstrate that the images in evening introduce more cross-domain discrepancy during learning. At the same time, the adaptation model performs worse because more training data provides more features during the transfer learning.

\begin{table}[]
\centering
\resizebox{\linewidth}{!}{%
\begin{tabular}{p{3.5cm}<{\centering}p{2cm}<{\centering}p{2.5cm}<{\centering}}
\toprule
Source Domain Dataset & ResNet-18 & ResNet-18-Adap \\ \midrule
All data & 72.24\% & \textbf{81.40\%} \\
\midrule
All winter data & 61\% & 68.57\% \\
All data without autumn & 64\% & 70.56\% \\
All morning data & \textbf{74.59\%} & 79.74\% \\ \bottomrule
\end{tabular}
}
\caption{Performance on Different Source Domain Dataset}\label{tb3}
\end{table}
%

\begin{table}[]
\centering
\resizebox{\linewidth}{!}{%
\begin{tabular}{p{5cm}<{\centering}p{3cm}<{\centering}}
\toprule
Source Domain Dataset & Use Adaptation \\ \midrule
All data & 81.40\% \\
\midrule
Four seasons & 82.28\%  \\
Morning and evening & \textbf{84.08\%} \\ \bottomrule
\end{tabular}
}
\caption{Multi-Source Adaptation Performance}
\label{multi}
\end{table}
%

\begin{figure}[]
\centering
\includegraphics[width=0.9\linewidth]{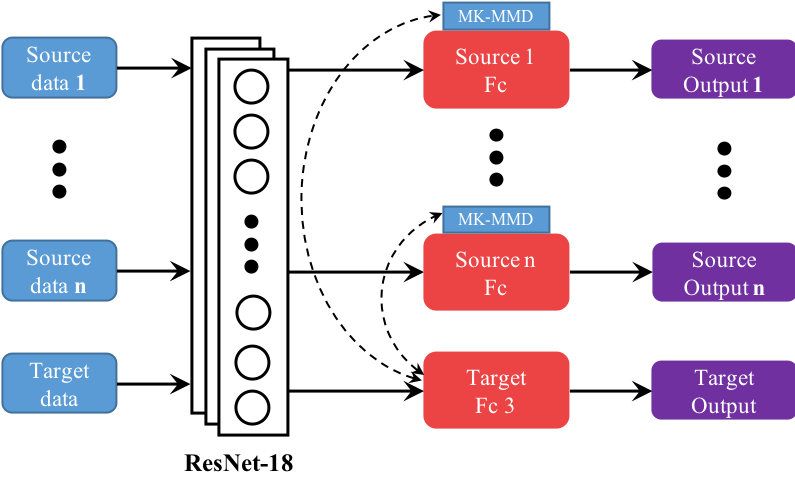}
\caption{Multi-Source adaptation network. For each source domain, images are fed into the network individually and the corresponding last fully connected layer is adapted by an MK-MMD layer between target domain. To simplify the problem, all source domains use the same learning weight $\lambda$.}
\label{multi_figure}
\end{figure}

\subsubsection{Multi-source adaptation}
From the experimental results above, we can make the following observations. (1) In all tasks, adaptation model achieves more than 80\% accuracy, which is well enough to make UAV fly automatically. (2) Different source data affects the adaptation performance to varying degrees. (3) The larger difference between domains, the worse performance achieves on basic model and the greater performance boost on adaptation model.

\begin{figure}[]
\centering
\includegraphics[width=0.8\linewidth]{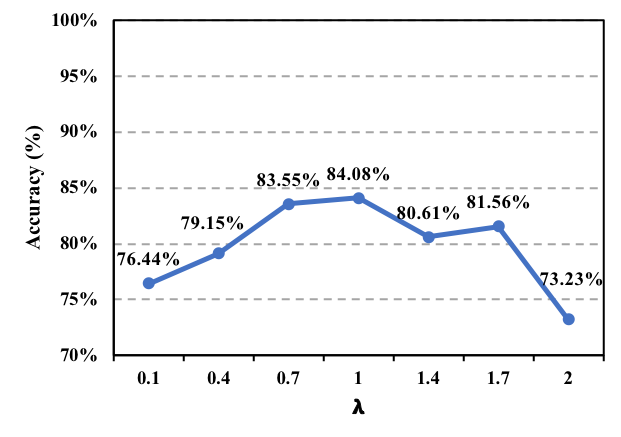}
\caption{Sensitivity of $\lambda$ based on multi light conditional adaptation network.}
\label{sensitivity}
\end{figure}

To dive deeper into domain adaptation, we construct a multi-source adaptation model based on ResNet-18 adaptation network to get fully use of different types of features extracted from each source domain (See Fig.\ref{multi_figure}).
The simulation dataset is separated based on seasonal and light condition difference (See Table \ref{multi}). In these two ways, the accuracies are both higher. Split by light condition, the multi-source adaptation model achieves 84.08\% accuracy, which is the best result.
Further, we make a sensitive test on learning weight of MK-MMD to investigate the effect of $\lambda$. From Fig.\ref{sensitivity}, we can observe that the accuracy first increases and then decreases. When $\lambda$ equals to 1, the best trade-off between learning forest features and adapting cross-domain discrepancy is achieved.

\section{Conclusion}
We formulate a classification problem to study autonomous navigating a monocular UAV in the dense forest. The navigating strategies are learned in simulation and transferred to the real world to avoid the obstacles and follow the trail. Our new pipeline saves the time of collecting data and reduces the risk of training a real aircraft. In this paper, we propose an adaptation network step by step, which achieves 84.08\% accuracy in the real-world test. The MK-MMD layer adapts the task-specific layers to jointly minimize the misclassification error and cross-domain discrepancies.
Inspired by the result that different types of data transfer specific features to target domain, we propose a simple multi-source adaptation network which achieves the best result.
What's more, we implement a simple flight controller and test it in the simulation environment.

One area of future work we plan to address is to construct an end-to-end multi-source adaptation network which can optimize the learning weight of each source domains automatically. In addition, a real flight test is on the schedule. Some engineering improvements should be considered in the future to make the flight stable.


%

\bibliographystyle{named}
\bibliography{tansferuav}

\end{document}